\documentclass{article}
\usepackage{ijcai26}

\usepackage{times}
\usepackage{soul}
\usepackage{url}
\usepackage[hidelinks]{hyperref}
\usepackage[utf8]{inputenc}
\usepackage[small]{caption}
\usepackage{graphicx}
\usepackage{amsmath}
\usepackage{amsthm}
\usepackage{booktabs}
\usepackage{amssymb}
\usepackage[switch]{lineno}

\usepackage{fancyhdr}
\usepackage{algorithm}
\usepackage{algpseudocode}
\usepackage{subfigure}
\usepackage{multirow}
\usepackage{subcaption}
\usepackage{cases}

\title{Beyond Fixed Patches: Enhancing GPTs for Financial Prediction with Adaptive Segmentation and Learnable Wavelets}

\author{
Renjun Jia$^{1*}$ \and
Zian Liu$^{2}$\footnote{Both authors contributed equally to this research}\and
Peng Zhu$^1$\and
Dawei Cheng$^{1}$\footnote{Corresponding author}\And
Yuqi Liang$^3$\\
\affiliations
$^1$School of Computer Science and Technology, Tongji University\\
$^2$School of Mathematical Sciences, Tongji University\\
$^3$Seek Data Group, Emoney Inc.\\
\emails
\{2332101, 2152632, pengzhu, dcheng\}@example.com,
roly.liang@seek-data.com
}
\begin{document}

\maketitle

\begin{abstract}
The extensive adoption of web technologies in the finance and investment sectors has led to an explosion of financial data, which contributes to the complexity of the forecasting task. Traditional machine learning models exhibit limitations in this forecasting task constrained by their restricted model capacity. Recent advances in Generative Pre-trained Transformers (GPTs), with their greatly expanded parameter spaces, demonstrate promising potential for modeling complex dependencies in temporal sequences. However, existing pretraining-based approaches typically focus on fixed-length patch analysis, ignoring market data's multi-scale pattern characteristics. In this study, we propose $\mathbf{GPT4FTS}$, a novel framework that enhances pretrained transformer capabilities for temporal sequence modeling through dynamic patch segmentation and learnable wavelet transform modules. Specifically, we first employ K-means++ clustering based on DTW distance to identify scale-invariant patterns in market data. Building upon pattern recognition results, we introduce adaptive patch segmentation that partitions temporal sequences while preserving pattern integrity. To accommodate time-varying frequency characteristics, we devise a dynamic wavelet transform module that emulates discrete wavelet transformation with enhanced flexibility in capturing time-frequency features. Extensive experiments on real-world financial datasets substantiate the framework's efficacy. The source code is available: \href{https://anonymous.4open.science/r/GPT4FTS-6BCC/}{https://anonymous.4open.science/r/GPT4FTS-6BCC/}.
\end{abstract}

\section{Introduction}

\begin{figure}[tbp]
    \centering
    
    \subfigure[CSI 300 Price]{
        \includegraphics[width=0.3\linewidth]{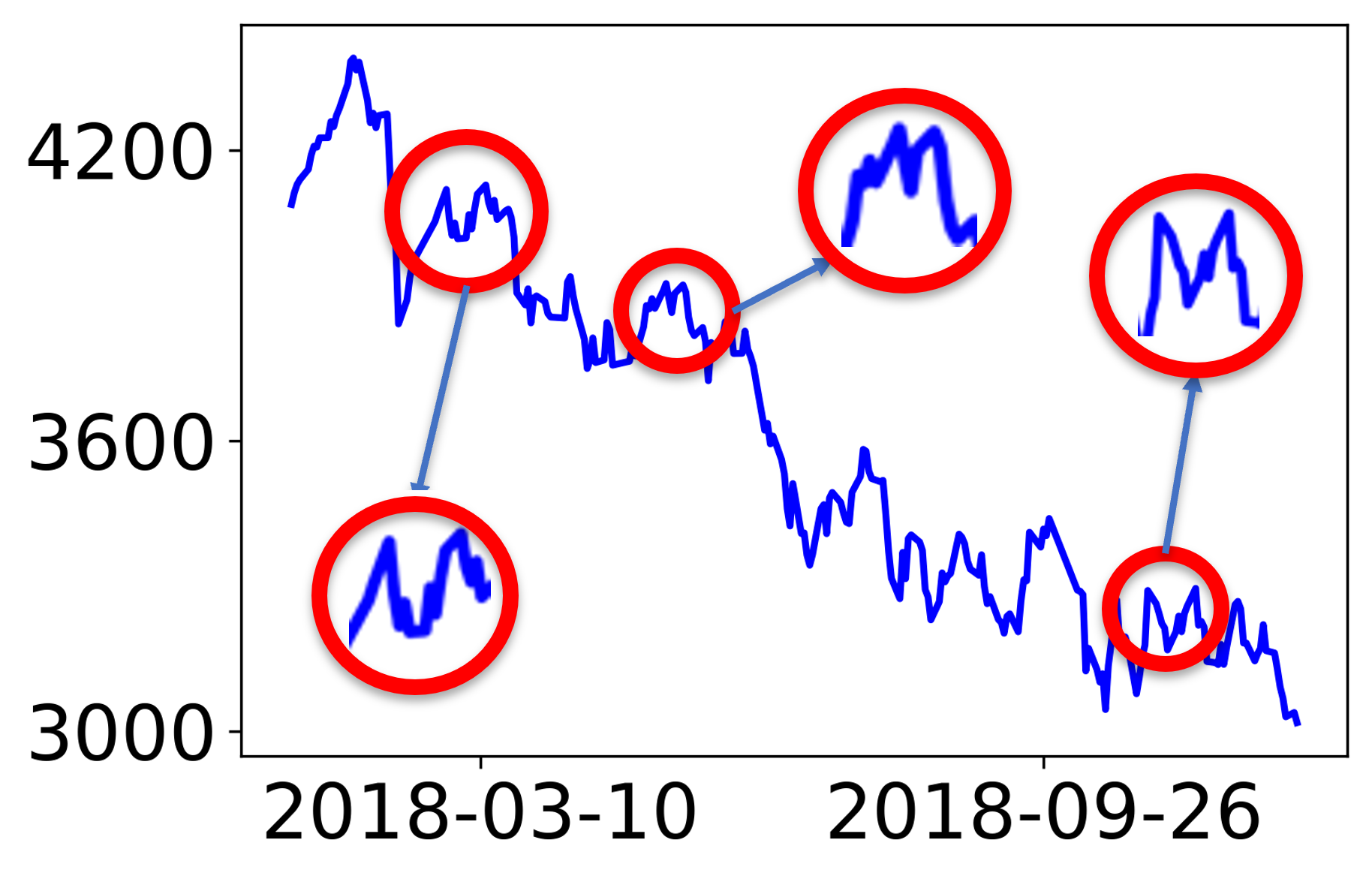}
    }
     \subfigure[Price Dist]{
        \includegraphics[width=0.3\linewidth]{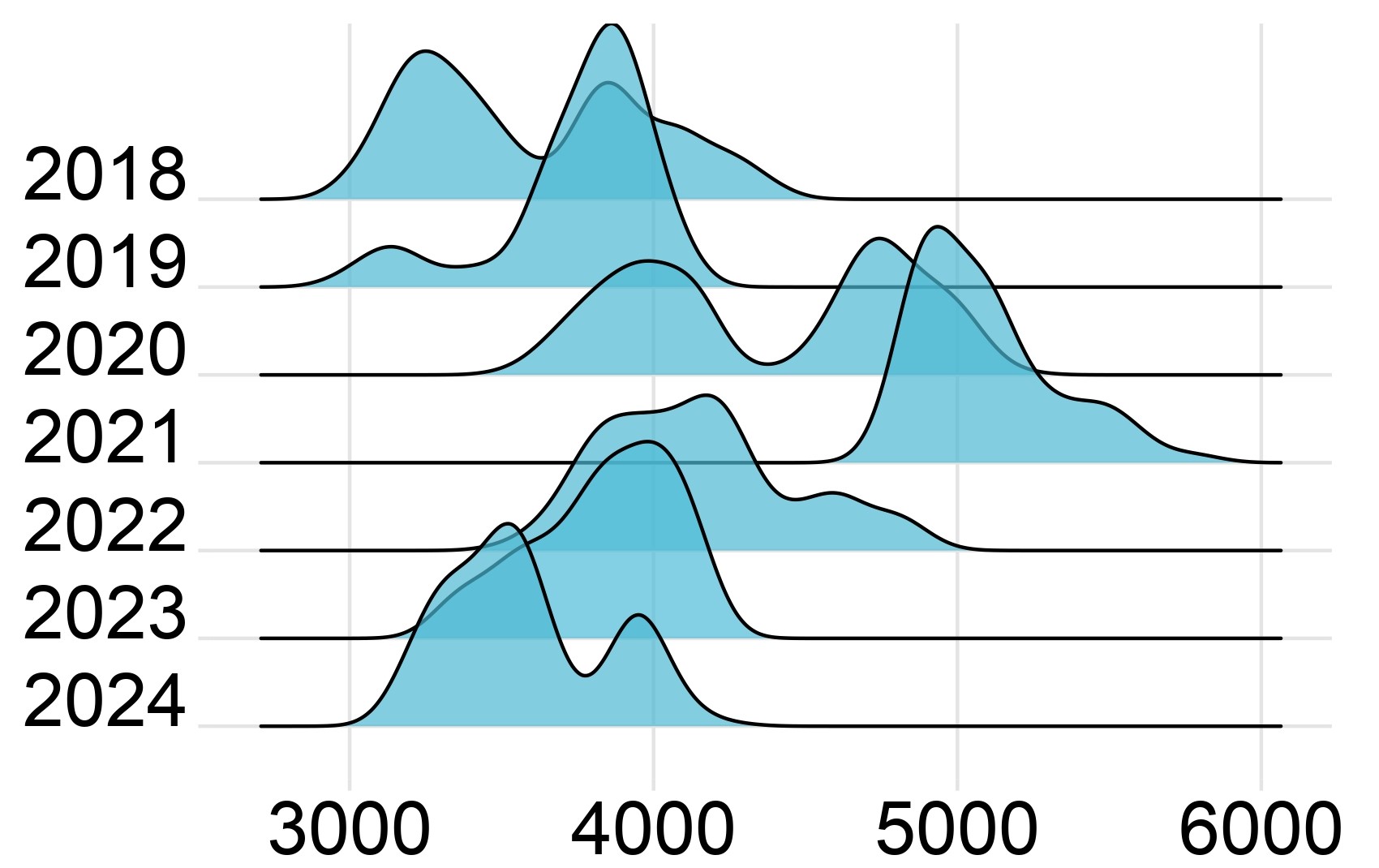}
    }
    \subfigure[Irregular Series]{
    \includegraphics[width=0.3\linewidth]{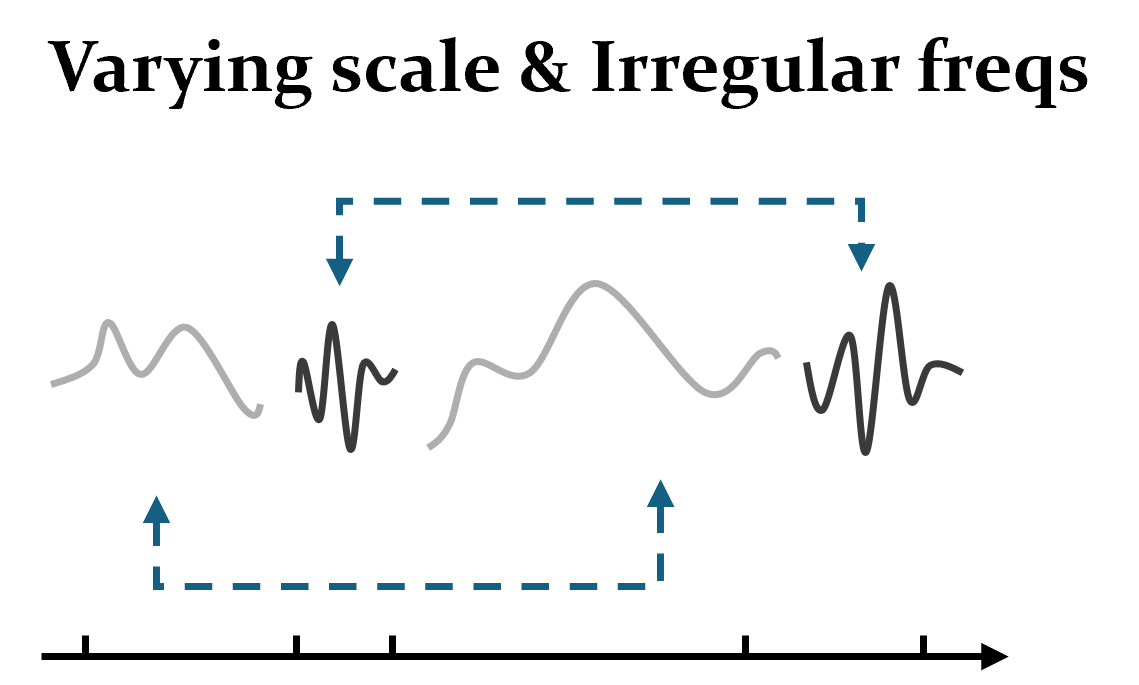}
    }
    \\
    \subfigure[Use of Electricity]{
	\includegraphics[width=0.3\linewidth]{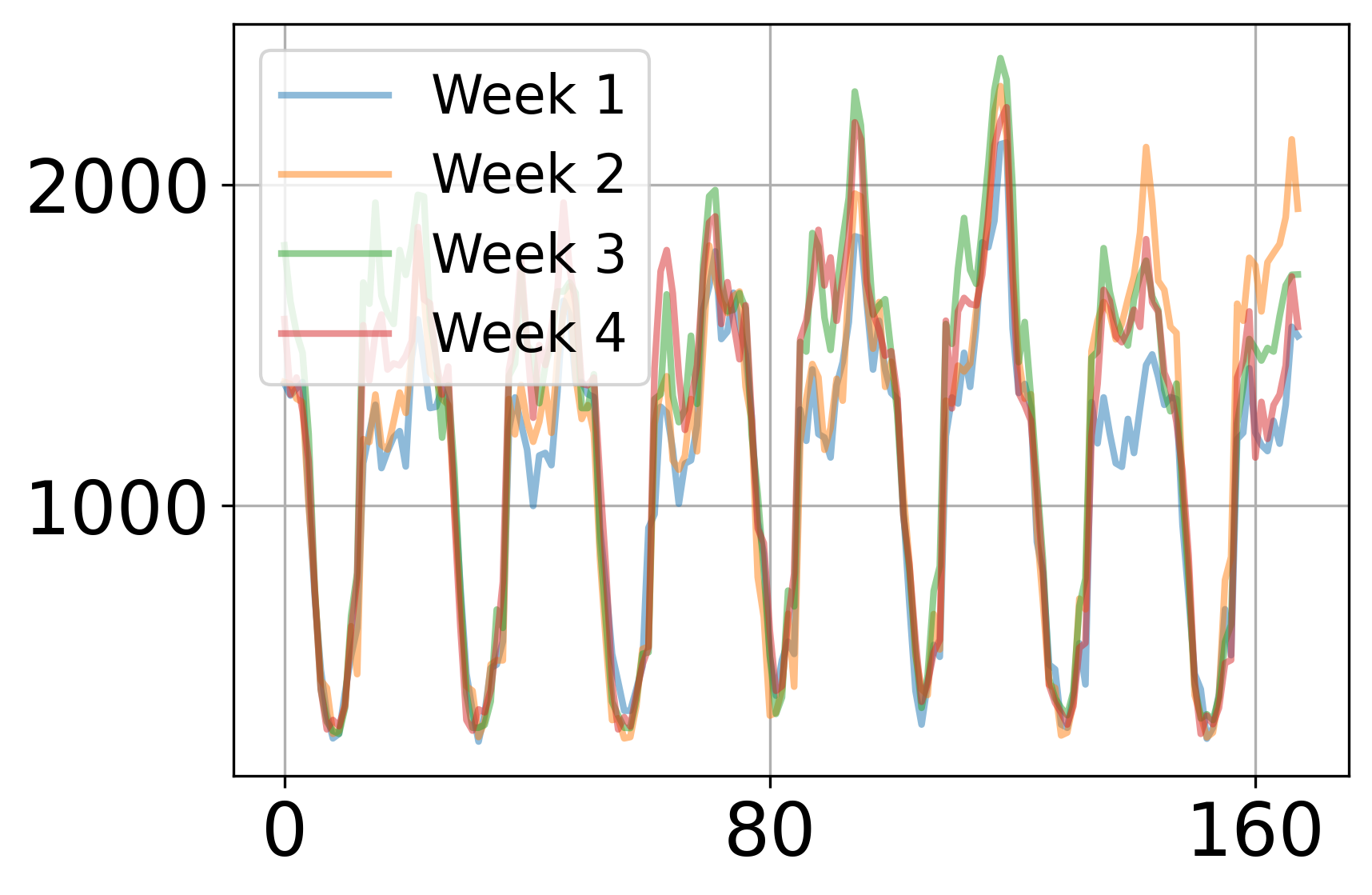}
    }
    \subfigure[Electricity Dist]{
	\includegraphics[width=0.3\linewidth]{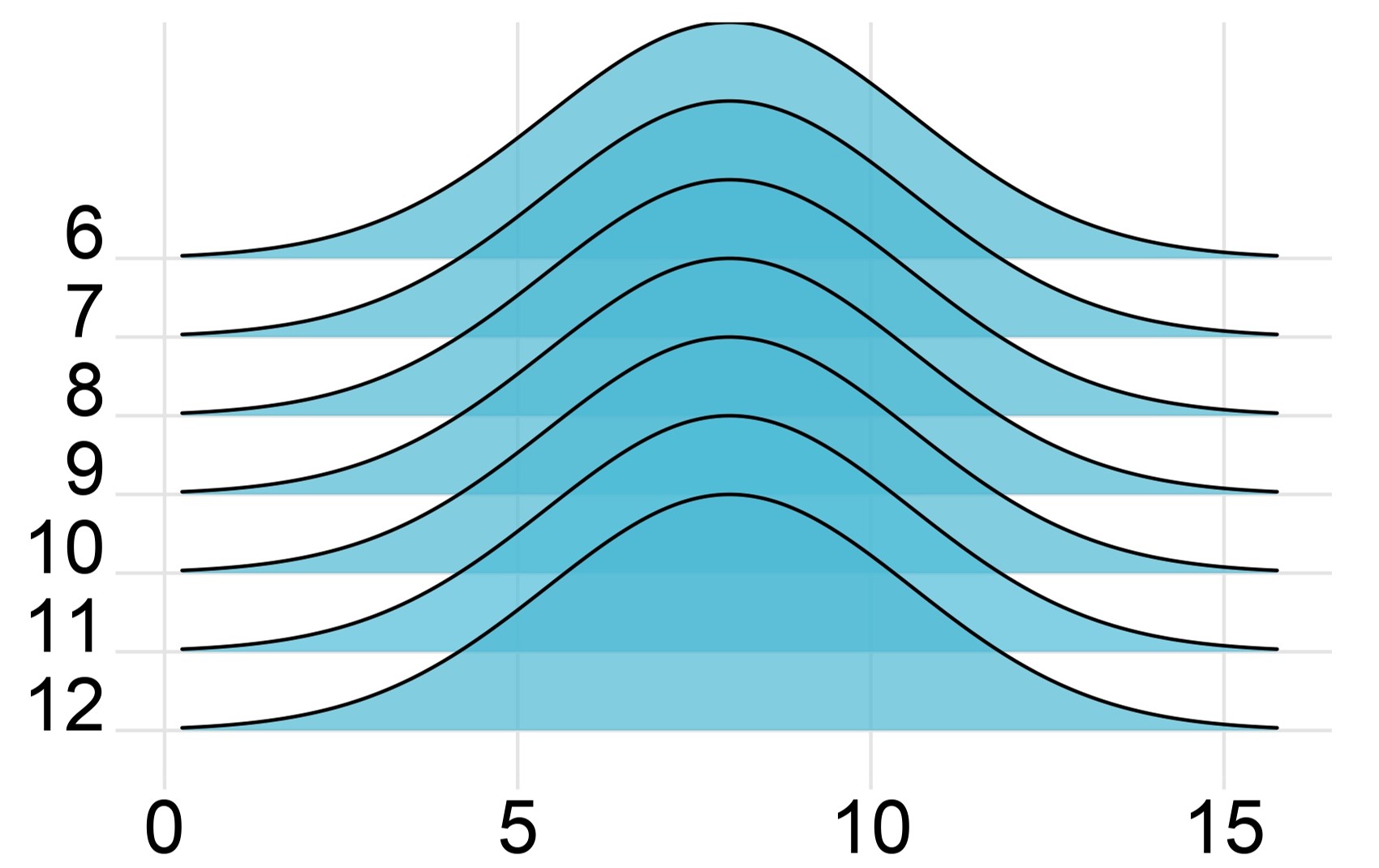}
    }
    \subfigure[Regular Series]{
    \includegraphics[width=0.3\linewidth]{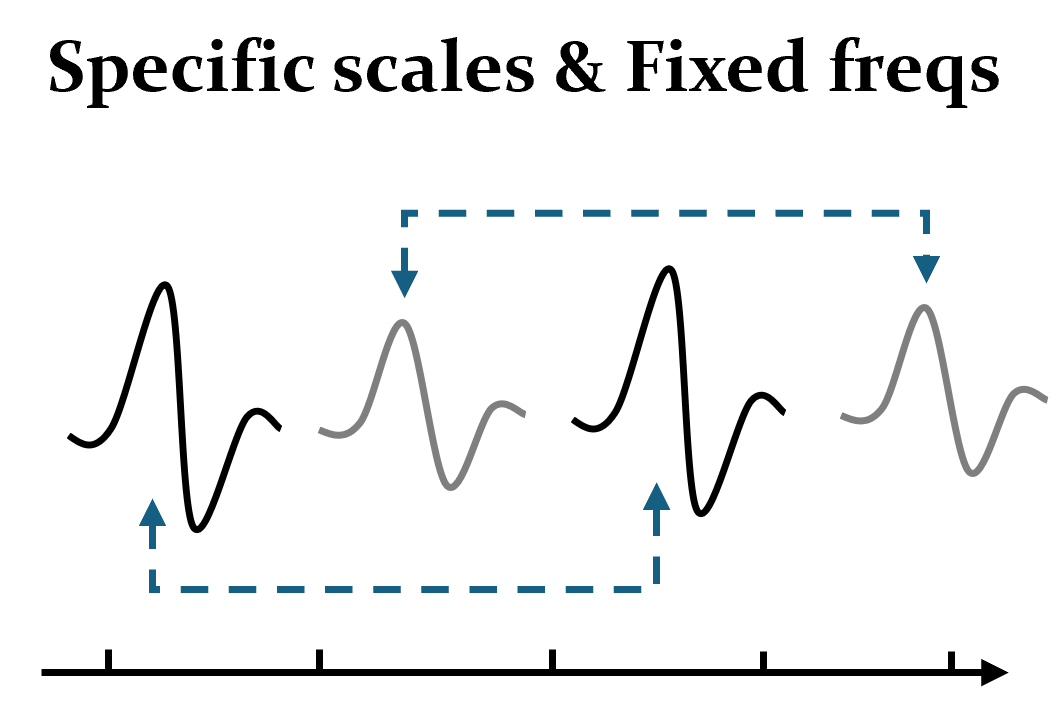}
    }
    \caption{Figures a and d reveal discrepancies in period patterns between financial and electricity data. Figures b and e illustrate deviations in their frequency domain distributions. Figures c and f abstract the differences.}
    \label{intro}
    \vspace{-1em}
\end{figure}

The task of financial time series prediction has grown profoundly challenging in quantitative analysis. While historically hindered by the inherently weak predictive signals obscured by market noise and non-stationary temporal dynamics \cite{huang2024generative}, this challenge is now significantly amplified by the data explosion facilitated by the widespread adoption of Web technology in finance, which introduces massive volumes of complex, high-velocity data that further complicate robust time series analysis. These challenges are aggravated by complex macro-economic interdependency\cite{ghironi2006macroeconomic}, irregular event-driven disturbance, and the heterogeneous behavior of market participants\cite{frijns2010behavioral}, which collectively manifest as intricate patterns that resist conventional market analysis methods. As illustrated in Figure \ref{intro}, the visual comparison between financial data and power data reveals that the intrinsic patterns in financial time series are more complex and exhibit similar patterns across different scales. Moreover, the frequency domain characteristics of financial time series vary over time, which contrasts sharply with power time series. All of the above highlights the low signal-to-noise ratio and complex dependencies characteristic of financial time series.
Traditional machine learning methods, including auto-regressive models\cite{taylor2016using} gradient-boosted decision trees\cite{machado2019lightgbm} and recurrent neural networks\cite{hochreiter1997long}, face limitations when applied to financial prediction tasks. These limitations stem from their restricted model capacity to simultaneously capture long-term sequence dependency, multi-resolution features, and nonlinear interactions among diverse market factors. Furthermore, the inherent assumption of stationarity in many traditional models fails to account for the non-stationary nature of financial time series, where statistical properties such as volatility and correlation structures evolve over time. For instance, while recurrent neural networks are capable of modeling sequences, their practical efficacy is undermined by gradient vanishing issues and inflexible receptive fields when processing extended financial sequences spanning multiple market regimes. This architectural rigidity prevents such models from adapting to the ever-evolving statistical property of financial markets.

Recent advances in enerative Pre-trained Transformers (GPTs) have demonstrated strong capabilities for time series modeling, leveraging their massive parameter scales to capture complex dependencies\cite{radford2019language,brown2020language,touvron2023llama}. The transformer's self-attention mechanism\cite{vaswani2017attention} has shown particular promise in modeling long-term dependencies through pairwise interactions across time steps\cite{wu2021autoformer,zhou2021informer}. In finance, initial applications reveal GPTs' potential in decoding structured temporal patterns for market prediction\cite{yu2023temporal,yang2023fingpt}, suggesting their adaptability beyond textual analysis to numerical forecasting tasks.

Current GPT-based financial prediction methods face critical limitations. For instance, Time-LLM \cite{jin2023time} uses textual prompting to reprogram time series, yet the complex modalities within financial patches are often poorly captured by simple textual prototypes. Moreover, most approaches rely on rigid, fixed-length patch segmentation \cite{bian2024multi,zhou2023one,cao2024tempo}, which arbitrarily divides time series without respecting the multi-scale nature of financial markets. This uniform segmentation disrupts semantically coherent patterns, discards contextual information, and breaks inherent temporal dependencies. Additionally, such static decomposition cannot adapt to the time-varying frequency characteristics of financial data, limiting model adaptability to evolving environments and structural breaks, ultimately impairing predictive performance.
% However, current LLM-based financial prediction methods exhibit critical flaws that limit their practical utility. For instance, Time-LLM \cite{jin2023time} employs textual information to reprogram time series, concatenating output prompts with reprogrammed patches to enrich the context. However, when dealing with complex financial time series, the modal information within each patch is challenging to accurately characterize through simple textual prototypes. Most existing methods rely on fixed-length temporal patch segmentation, arbitrarily dividing time series into uniform intervals without regard for the intricate multi-scale nature of market reality\cite{bian2024multi,zhou2023one,cao2024tempo}. This uniform segmentation inevitably breaks semantically coherent patterns, thereby discarding critical contextual information and disrupting the inherent temporal dependencies essential for accurate forecasting. Furthermore, the static nature of conventional patch decomposition fails to accommodate the time-varying frequency characteristics of financial data, where different frequency components dynamically influence price movements. This rigidity fundamentally constrains the model's adaptability to evolving market conditions and structural breaks, ultimately compromising its predictive performance in real-world trading scenarios.

Recognizing the potential of GPTs in financial time series modeling and the limitations of existing approaches, we propose the GPT4FTS framework to fully realize the potential of GPTs for financial time series prediction. Our framework integrates an offline scale-invariant pattern recognition algorithm\cite{huang2024generative}, a learnable patch segmentation strategy, and a dynamic wavelet transform module \cite{chen2025simpletm} to enhance the modeling of financial time series data by capturing its multi-scale characteristics and complex temporal dependencies. In summary, the primary contributions of this work include:
\begin{itemize}
\item To the best of our knowledge, this is the first work to enhance foundation language models to model the complex interactions between patches while capturing the scale-invariant patterns in financial time series.

\item  We devise scale-invariant pattern recognition algorithm, learnable patch segmentation strategy, and dynamic wavelet transform module to collectively enhance GPTs' capability in modeling financial time series.

\item We conduct experiments on four real-world datasets, showing that GPT4FTS outperforms state-of-the-art baselines. Ablation studies confirm improved accuracy.

\end{itemize}

\begin{figure*}[tbp]
    \centerline{\includegraphics[width=\linewidth]{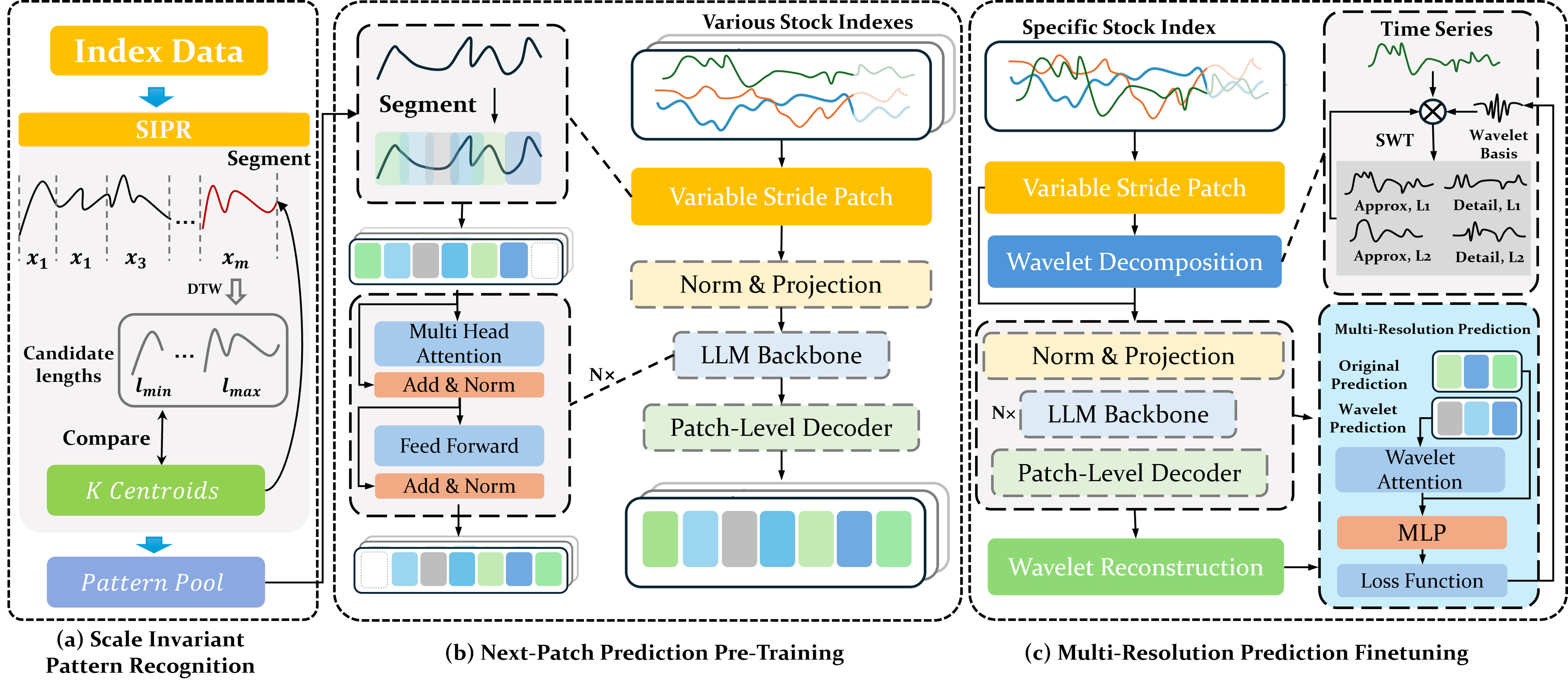}}
	\caption{The architecture of the proposed GPT4FTS model. Part (a) outlines the off-line learning time series scale invariant pattern recognition module, which uses k-means++ clustering and DTW algorithms to match patterns of sequences of different lengths to obtain sequence cuts that minimize the total distance. Part (b) describes the training process using dynamic stepping. Part (c) introduces a learnable wavelet transform module that adaptively decomposes the input sequences.}
	\label{fig:framework}
\end{figure*}
\section{Related Works}

\paragraph{Methods For Financial Prediction}
Financial prediction has evolved through several methodological shifts. Traditional approaches, including Exponential Smoothing \cite{de2009predicting} and ARIMA \cite{ariyo2014stock}, were valued for modeling linear trends. Subsequently, machine learning techniques like Support Vector Machines \cite{kim2003financial} and XGBoost \cite{wang2020forecasting} became prominent with increased data and computational resources. However, the high volatility and low signal-to-noise ratio of financial data often lead these models to overfit. Deep learning introduced Recurrent Neural Networks (RNNs) to better capture sequential dependencies \cite{di2017recurrent}, though their ability to model long-range relationships remains limited. To address this, methods like diffusion models have been adapted from image generation to synthesize financial time series and capture complex patterns, as seen in FTS-Diffusion \cite{huang2024generative}.More recently, Transformers \cite{wu2021autoformer,zhou2021informer} have gained traction for their multi-head self-attention and parallel computation, which efficiently handle long sequences. And growing body of work has begun to focus on the inherent properties of the market itself \cite{Yang2025LearningUM}.

% Despite their advantages, the noisy nature of financial data often necessitates integration with noise reduction or information enhancement techniques \cite{zeng2023financial, olorunnimbe2024ensemble}.
\paragraph{GPTs for Financial Prediction}
Building on the success of Transformers, Generative Pre-trained Transformers (GPTs) have demonstrated remarkable capabilities not only in NLP and CV~\cite{zhu2023minigpt,brown2020language,touvron2023llama} but also show promise for financial time series modeling. Their profound capacity for contextual understanding and knowledge reasoning is particularly suited to tackling the complex, noisy patterns characteristic of financial data \cite{jin2024position,tang2024time}. Existing GPT-based financial prediction methods fall into two main categories. The first category involves training GPTs with market-driven feedback. For instance, some works \cite{wang2024stocktime,yu2023temporal} leverage textual news and price data for forecasting by treating prices as token sequences or using GPTs' inherent reasoning capabilities. A common limitation of these approaches is their reliance on external, high-quality text data, which can be difficult to obtain. The second category adapts general-purpose GPTs for time series through prompting, reprogramming, or fine-tuning \cite{jin2023time,cao2024tempo,zhou2023one,bian2024multi}. While effective, these methods often employ fixed-length patch strategies that overlook the multi-scale patterns inherent in financial data. Unlike these approaches, our framework dynamically adapts to different scales through variable-length patches, explicitly capturing the intrinsic multi-scale characteristics and cross-scale dependencies in financial time series. This distinguishes our work from both general-purpose time-series models \cite{liutimer,liu2024moiraimoe} and existing financial GPTs, providing a tailored solution for the unique challenges of financial forecasting.

\section{Methodology}

\subsection{Problem Formulation}
The definition of the stock forecasting problem can vary depending on the specific investment strategy adopted by the investor. In this paper, we adopt a paradigm widely accepted in the research field, namely cross-sectional analysis\cite{Asgharian2002CrossSA}. Building on existing work, we input historical data on normalized stocks with multiple metrics and output a predict of the next day's return score. Given a set of $B$ stocks $X = \{x_1, x_2, ..., x_n\}$ consisting of all data from the stock market, each stock $x_i \in \mathbb{R}^{L \times M}$  contains historical data with a backtracking window length $L$ , where $M$ denotes the indicator dimension at one time step. Our task is to predict the return $r^t_i$ on trading day $t$. Denoting our model parameters as $\Theta$, the process can be expressed as follows:
\begin{equation*}
    X \in \mathbb{R}^{B\times M\times L} \stackrel{\Theta}{\longrightarrow} r \in \mathbb{R}^{B}
\end{equation*}
In this paper, we focus on predicting the stock's return the next day. Therefore, the label of the predicted value r is defined as $r^s_t = p^s_{t+1} / p^s_{t} -1  $, where $p^s_t$ denotes the closing price of the stock.

\subsection{Scale-Invariant Pattern Recognition}
In this section, we propose a method for identifying patterns in financial time-series data based on the scale-invariant property inherent to such data. A suitable approach involves employing clustering techniques for sequence model recognition. While investigating the impact of different clustering methods on subsequent predictions remains an important research direction, a detailed discussion of such methodological nuances falls beyond the scope of this study. For the purpose of this work, we focus on the K-means algorithm to establish a robust baseline.

Specifically, we partition the entire financial time-series into variable-length segments and group them into K distinct clusters using the fundamental K-means algorithm with Dynamic Time Warping (DTW) as the distance metric. Unlike conventional approaches that apply fixed-length partitioning across entire time series, we adopt a simple yet effective splitting method to determine the optimal segment length for each portion adaptively. For the candidate length $l \in [l_{min},l_{max}]$ at position $t = \sum_{\tau=0}^{m-1}t_{\tau}$ of the sequence, we obtain it by minimizing the distance between the subsequence and the cluster centroids pattern for each possible length, while $x_m=X_{t:t+l^*}$ is considered as the optimal segmentation:
\begin{equation}
    l^*= \mathop {arg\min}_{l \in [l_{min},l_{max}]}d(X_{t:t+l},\mathbf{p}),\forall \mathbf{p} \in \mathcal{P}
\end{equation}
where $m$ is the number of sequences that have been segmented and $\mathbf{p}$ is the cluster centroid sequence used for the comparison.

To address the need for calculating distances between segments of varying lengths and to focus on the differences between various modes, we adopt the DTW distance measure in place of the standard Euclidean distance method. Consider $X = \{x_1, x_2, \dots, x_n\}$ and $Y = \{y_1, y_2, \dots, y_m\}$, where $n$ and $m$ denote the lengths of sequences $X$ and $Y$, respectively. The DTW algorithm computes the minimum cumulative distance between these sequences, allowing for non-linear alignments.
The cumulative distance $D_{i,j}$ is defined as:
\begin{equation}
    D_{i,j} = d(x_i, y_j) + \min(D_{i-1,j},D_{i,j-1},D_{i-1,j-1})
\end{equation}
where $d(x_i, y_j)$ is the local distance between points $x_i$ and $y_j$, typically the Euclidean distance, although it can be adapted to other metrics. The final DTW distance is $D_{N,M}$, representing the optimal alignment cost between the sequences.

Inspired by the method used in FTS-Diffusion\cite{huang2024generative}, we adapt it by using volatility-based weighted DTW distances in financial time series to mitigate the adverse effects of random initialization on clustering results. Specifically, we first randomly select a segment of the predetermined length from all available segments. Then, from the remaining segments, we choose the one that is farthest from the already selected segment to ensure sufficient dissimilarity among the centroids. This method demonstrates greater robustness when dealing with segments of varying shapes.

\subsection{Next-patch Prediction Pre-Training}
In this section, we propose using next-patch prediction as a continual pre-training task. We introduce an efficient dynamic patch partitioning strategy that adapts to historical sequence segmentation based on references learned from market index data. This method guides partitioning stock data across temporal phases while maintaining training efficiency.

Following \cite{bian2024multi}, we frame time-series forecasting as an autoregressive output process of the language model, enabling it to understand time-series patches. 

Given time-series data, we flatten them into $M$ univariate sequences, where the $i$-th sequence with a look-back window size $L$ starting at time $t$ is denoted as ${x_t^i, \dots, x_{t+L-1}^i} \in \mathbb{R}^L$. Each sequence is then divided into overlapping patches ${p_{t_p}^i, \dots, p_{t_p+L_p-1}^i} \in \mathbb{R}^{L_p \times P}$, with patch extraction dynamically determined by segment positions and a sliding window strategy. For each batch, we adjust segment positions relative to the batch start and combine them with stride-based starts to form the complete set of patch extraction points.

\subsection{Multi-Resolution Prediction Fine-tuning}
\begin{figure}[tbp]
 \includegraphics[width=1\linewidth]{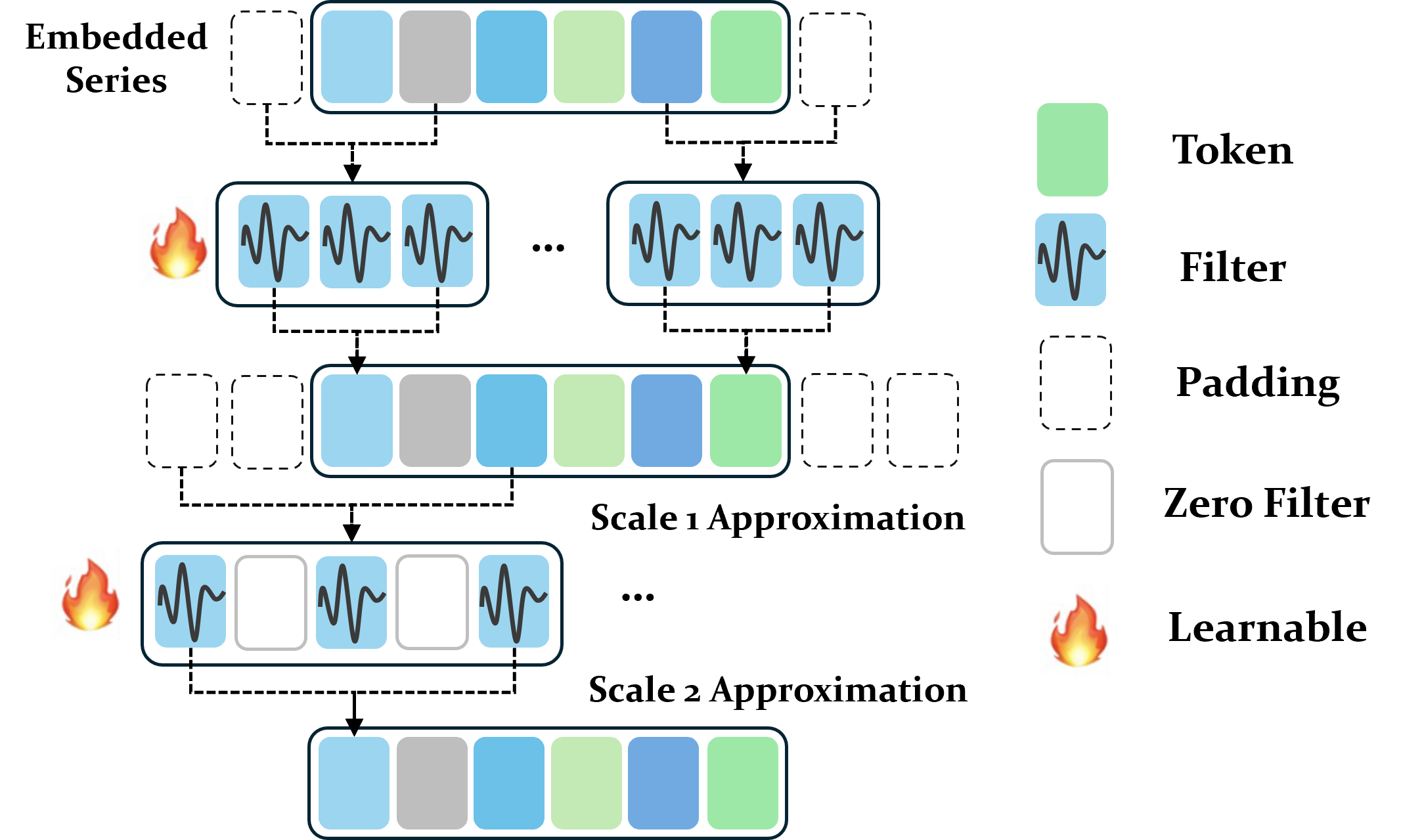}
\caption{The SWT tokenization method employs input padding and incorporates learnable filters with zero-insertion operations.}
\vspace{-1em}
\end{figure}
This section proposes a tokenization framework for financial time series. The method extracts multi-scale temporal features, from high-frequency fluctuations to low-frequency trends, while preserving both local patterns and global characteristics of each variable. This comprehensive representation aids the integration of temporal information for forecasting. Scale-specific processing modules can further enhance predictive performance by handling different temporal resolutions separately.

The wavelet transform naturally meets these requirements, as demonstrated in transformer-based vision architectures \cite{wavevit2022}. It decomposes signals across scales while maintaining precise time localization. Our framework processes each wavelet scale independently to capture scale-dependent feature interactions, filtering information relevant to tasks where predictive dependencies are scale-specific. To address financial non-stationarity, we employ a trainable wavelet transform that adapts to market dynamics.

Data from different stocks are processed as independent channels. A learnable Stationary Wavelet Transform (SWT) generates a multi-scale representation:
\begin{equation}
SWT(\cdot;h_0,g_0):\mathbb{R}^{M\times L}\rightarrow \mathbb{R}^{M\times L \times (S+1)},
\end{equation}
using learnable filters \(h_0, g_0 \in \mathbb{R}^{M \times k}\) of kernel size \(k\) across \(S\) decomposition levels. It produces time-frequency tokens \(\{t_1^{(s)},\dots,t_L^{(s)}\}_{s=0}^S\) that retain the original length \(L\) via zero-padding and no downsampling, ensuring time-invariance across the sequence. This scale-invariant, channel-wise processing captures localized events across different temporal scales while keeping variables independent.

The transform is mathematically founded on a mother wavelet \(\psi(t)\), which encodes localized oscillations, and a scaling function \(\phi(t)\), which captures smooth approximations. From these, a family of discrete wavelet basis functions is derived through scaling and translation:
\begin{subnumcases}{}
\psi_{s,k}(t) = 2^{-s/2} \psi(2^{-s}t-k) \label{eq:wavelet_basis} \\
\phi_{s,k}(t) = 2^{-s/2} \phi(2^{-s}t-k) \label{eq:scaling_basis}
\end{subnumcases}
Here, \(s\) is the scale parameter (larger \(s\) corresponds to lower frequency, broader support), and \(k\) is the translation parameter. The factor \(2^{-s/2}\) ensures energy normalization across scales. Equation \eqref{eq:wavelet_basis} generates wavelet functions that extract detail (high-frequency) information, while Equation \eqref{eq:scaling_basis} generates scaling functions that extract approximation (low-frequency) information.

The transform decomposes the input series \(\{x_t\}^{L}_{t=1}\) using a low-pass filter \(h\) (from \(\phi(t)\)) and a high-pass filter \(g\) (from \(\psi(t)\)):
\begin{subnumcases}{}
h(k)=\langle \phi(t),\phi(2t-k) \rangle \label{eq:lowpass_filter} \\
g(k)=\langle \psi(t),\phi(2t-k) \rangle \label{eq:highpass_filter}
\end{subnumcases}
Filter \(h\) in \eqref{eq:lowpass_filter} captures smooth trends; filter \(g\) in \eqref{eq:highpass_filter} extracts high-frequency details. Starting with \(c_t^{(0)} = x_t\), the decomposition at level \(s\) is:
\begin{subnumcases}{}
c_t^{(s+1)}=\sum h^{(s)}(k) c_{t+k}^{s} \label{eq:approx_coeff} \\
d_t^{(s+1)}=\sum g^{(s)}(k) c_{t+k}^{s} \label{eq:detail_coeff}
\end{subnumcases}
where \(c_t^{(s)}\) are approximation coefficients (low-frequency trends) and \(d_t^{(s)}\) are detail coefficients (high-frequency components). Filters \(h^{(s)}\) and \(g^{(s)}\) are upsampled by inserting \(2^s-1\) zeros to preserve length \(L\). Unlike fixed filters, \(h\) and \(g\) are learnable parameters \cite{chen2025simpletm,michau2022}, allowing the decomposition to adapt and optimize for discriminative patterns in each variate, merging wavelet theory with data-driven flexibility.

\begin{table*}[t]
\small
\caption{Comparing the experimental results of the models on four datasets. ARR measures the portfolio return rate of each predictive model, with higher values being better. AVol and MDD measure the investment risk of each predictive model, with lower absolute values being better. ASR, CR, and IR measure profits under unit risk, with higher values being better.}
% We take 5 different random seeds for each model for training and take the best of them to record the results.
\setlength{\tabcolsep}{7.5pt}
\centerline{
\begin{tabular}{lcccccccccccc}

\toprule
Datasets     & \multicolumn{6}{c}{CSI300} & \multicolumn{6}{c}{CSI500}  \\
\cmidrule(l){2-7} \cmidrule(l){8-13}
Model        & ARR$\uparrow$  &  AVol$\downarrow$  & MDD$\downarrow$   & ASR$\uparrow$   & CR$\uparrow$   & IR$\uparrow$    & ARR$\uparrow$  &  AVol$\downarrow$  & MDD$\downarrow$   & ASR$\uparrow$   & CR$\uparrow$   & IR$\uparrow$    
\\ 

\midrule
LSTM         & 0.104 & 0.243 & 0.173 & 0.431 & 0.605 & 0.536 & 0.161 & 0.313 & 0.199 & 0.514 & 0.808 & 0.656 \\
GRU          & 0.166 & 0.234 & 0.154 & 0.707 & 1.076 & 0.779 & 0.135 & 0.292 & 0.199 & 0.461 & 0.677 & 0.565 \\
Transformer  & 0.235 & 0.221 & 0.158 & 1.065 & 1.492 & 1.112 & 0.193 & 0.306 & 0.228 & 0.629 & 0.845 & 0.695 \\
% ALSTM        & 0.287 & \textbf{0.198} & \textbf{0.073} & 1.444 & \underline{3.904} & 1.442 & 0.240 & 0.310 & 0.279 & 0.775 & 0.861 & 0.866 \\ 
AlphaStcok  & 0.308& \underline{0.215} & \textbf{0.105}& 1.431& \underline{2.924} & 1.360& 0.051&  \underline{0.273}& 0.172& 0.187& 0.297& 0.318 \\
DeepPocket  & 0.207 & \textbf{0.203} & 0.135 & 1.016 & 1.528 & 1.029 & 0.141 & \textbf{0.260} & 0.174 & 0.541 & 0.809 & 0.637 \\
DeepTrader   & \underline{0.385} & 0.293 & 0.162 & 1.313 & 2.377 & 1.323 & 0.273 & 0.331 & \underline{0.155} & 0.825 & 1.759 & 1.002 \\
MASTER & 0.194 & 0.223 & \underline{0.107} & 0.869 & 1.816 & 0.960 & 0.413 & 0.333 &0.205 & 1.241 & 2.013 & 1.201 \\ 
UMI   & 0.297  & 0.237  & 0.131  & 1.262  & 2.277  & 0.077  & 0.287  & 0.262 & 0.193  & 1.095 & 1.484& 0.069\\
Dlinear    & 0.192 & 0.287 & 0.143 & 0.669 & 1.341 & 0.816 & 0.347 & 0.336 & 0.174 & 1.033 & 1.987 & 1.070 \\
PatchTST  & 0.308 & 0.243 & 0.141 & 1.265 & 2.174 & 1.213 &  0.245 &  0.281 &\textbf{0.129} &  0.872  & 1.903 & 0.875 \\
iTransformer & 0.372 & 0.309 & 0.148 & 1.203 & 2.498 & 1.184 & 0.218 & 0.329 & 0.149 & 0.662 & 1.461 & 0.748 \\
Crossformer  & 0.359 & 0.234 & 0.157 & \underline{1.532} & 2.280 & \underline{1.520} & 0.307 & 0.296 & 0.187 & 1.034 & 1.635 & 1.016 \\ 
TimeMixer & 0.395 & 0.272 & 0.172 & 1.467 & 2.560 & 1.324 & 0.165 & 0.343 & 0.246 & 0.481 & 0.671 & 0.583 \\
GPT4TS       & 0.333 & 0.330 & 0.198 & 1.009 & 1.682 & 1.103 & 0.519 & 0.344 & 0.200 & 1.510 & 2.590 & 1.378 \\
TIME-LLM     & 0.370 & 0.323 & 0.209 & 1.145 & 1.771 & 1.205 & \underline{0.563} & 0.340 & 0.194 &  \underline{1.704} & \textbf{2.990} & \underline{1.521} \\
aLLM4TS      & 0.312 & 0.331 & 0.177 & 0.943 & 1.764 & 1.057 & 0.376 & 0.337 & 0.247 & 1.115 & 1.523 & 1.115 \\
\textbf{GPT4FTS}         & \textbf{0.528} & 0.233 & 0.109 & \textbf{2.262} & \textbf{4.808} & \textbf{1.965} & \textbf{0.643} & 0.351 & 0.226 & \textbf{1.829} & \underline{2.839} & \textbf{1.591} \\ 
\midrule
Datasets     & \multicolumn{6}{c}{S\&P500} & \multicolumn{6}{c}{NDX100} \\
\cmidrule(l){2-7} \cmidrule(l){8-13}
Model        & ARR$\uparrow$  &  AVol$\downarrow$  & MDD$\downarrow$   & ASR$\uparrow$   & CR$\uparrow$   & IR$\uparrow$    & ARR$\uparrow$  &  AVol$\downarrow$  & MDD$\downarrow$   & ASR$\uparrow$   & CR$\uparrow$   & IR$\uparrow$     \\ \midrule
LSTM         & 0.183 & 0.126 & 0.070 & 1.450 & 2.611 & 1.416  & 0.140 & 0.165 & 0.095 & 0.852 & 1.470 & 0.883 \\
GRU          & 0.204 & 0.131 & 0.075 & 1.558 & 2.697 & 1.456 & 0.229 & 0.239 & 0.148 & 0.957 & 1.542 & 1.088 \\
Transformer  & 0.244 & 0.145 & 0.102 & 1.682 & 2.376 & 1.630  & \underline{0.258} & 0.271 & 0.221 & 0.951 & 1.167 & 1.074 \\
% ALSTM        & 0.236 & 0.151 & 0.103 & 1.556 & 2.281 & 1.546  & 0.235 & 0.235 & 0.172 & 0.918 & 1.370 & 1.070 \\ 
AlphaStcok  & 0.148& 0.118& \underline{0.057}& 1.257& 2.584& 1.236& 0.131& 0.172& 0.123& 0.759& 1.065& 0.803  \\
DeepPocket  & 0.134 & \textbf{0.116} & 0.065 & 1.156 & 2.056 & 1.147 & 0.106 & \textbf{0.145} & 0.097 & 0.732 & 1.099 & 0.771  \\
DeepTrader   & 0.171 & \underline{0.118} &\textbf{0.049} & 1.457 & 3.467 & 1.460  & 0.183 & 0.196 & 0.108 & 0.934 & 1.698 & 1.161 \\
MASTER & 0.150 & 0.147 & 0.079 & 1.014 & 1.896 & 1.032 & 0.229 & 0.194 & 0.151 & 1.180 & 1.515 & 1.219 \\ 
UMI   & 0.086  & 0.126  & 0.078  & 0.679  & 1.092  & 0.045  &  0.076  & 0.111 & 0.066  & 0.686 & 1.144 & 0.045\\
Dlinear    & 0.167 & 0.150 & 0.085 & 1.111 & 1.952 & 1.095  & 0.081 & 0.222 & 0.181 & 0.362 & 0.444 & 0.489 \\
PatchTST  & 0.176&  0.166  & 0.087  &  1.063  &   2.024&   1.089  & 0.206   &  0.173  &  0.112  & \underline{1.190}   & 1.844  &  \underline{1.279}   \\
iTransformer & 0.082 & 0.156 & 0.095 & 0.523 & 0.860 & 0.644  & 0.088 & 0.253 & 0.163 & 0.349 & 0.544 & 0.590 \\
Crossformer  & 0.228 & 0.141 & 0.088 & 1.613 & 2.600 & 1.537  & 0.192 & 0.231 & 0.128 & 0.831 & 1.498 & 0.944 \\ 
TimeMixer & 0.048 & 0.149 & 0.105 & 0.323 & 0.458& 0.454 & 0.103 & 0.248 & 0.223 & 0.417 & 0.463 & 0.578 \\
GPT4TS       & \underline{0.321} & 0.157 & 0.073 & \underline{2.034} & \underline{4.346} & \underline{1.872}  & 0.242 & 0.221 & \underline{0.077} & 1.093 & \underline{3.116} & 1.104 \\
TIME-LLM     & 0.130 & 0.240 & 0.155 & 0.543 & 0.842 & 0.682 & 0.183 & 0.246 & 0.139 & 0.745 & 1.320 & 0.896 \\
aLLM4TS      & 0.236 & 0.159 & 0.083 & 1.481 & 2.821 & 1.396  & 0.183 & 0.210 & 0.119 & 0.869& 1.538&	0.879\\
\textbf{GPT4FTS}         & \textbf{0.371} & 0.145 & 0.081 & \textbf{2.559} & \textbf{4.585} & \textbf{2.258} & \textbf{0.446} & \underline{0.161} & \textbf{0.063} & \textbf{2.772} & \textbf{7.066} & \textbf{2.371} \\ \bottomrule
\end{tabular}}
\label{tab:port}
\end{table*}

\begin{table*}[tbp]
\small
\centering
\caption{Comparison of MSE ($10^{-3}$) and MAE ($10^{-3}$) metrics across four financial time series datasets using different methods.}
\setlength\tabcolsep{4pt}
\begin{tabular}{c|c|ccccccccccc}
\toprule
Dataset & Metric & LSTM & TimeMixer & MASTER & Dlinear & PatchTST & iTrans & GPT4TS & Time-LLM & aLLM4TS & \textbf{GPT4FTS} \\
\midrule
\multirow{2}{*}{CSI 300} & MSE & 0.487 & 0.498 & 0.491 & 0.484 & 0.489 & 0.511 & 0.575 & 0.550 & 0.491 & \textbf{0.475} \\
                          & MAE  & 1.487 & 1.522 & 1.496 & 1.485 & 1.509 & 1.562 & 1.659 & 1.618 & 1.534 & \textbf{1.469} \\
\midrule
\multirow{2}{*}{CSI 500} & MSE & 0.726 & 0.742 & 0.731 & 0.713 & 0.728 & 0.767 & 0.811 & 0.732 & 0.799 & \textbf{0.709} \\
                          & MAE  & 1.898 & 1.930 & 1.910 & 1.887 & 1.908 & 1.990 & 2.168 & 1.912 & 2.014 & \textbf{1.876} \\
\midrule
\multirow{2}{*}{NDX 100}   & MSE & 0.484 & 0.496 & 0.479 & 0.476 & 0.481 & 0.511 & 0.495 & 0.483 & 0.465 & \textbf{0.461} \\
                          & MAE  & 1.459 & 1.483 & 1.452 & 1.442 & 1.446 & 1.529 & 1.709 & 1.453 & 1.416 & \textbf{1.404} \\
\midrule
\multirow{2}{*}{S\&P 500}   & MSE & 0.338 & 0.345 & 0.339 & 0.332 & 0.337 & 0.342 & 0.348 & 0.335 & 0.330 & \textbf{0.321} \\
                          & MAE  & 1.232 & 1.247 & 1.239 & 1.216 & 1.225 & 1.245 & 1.256 & 1.220 & 1.212 & \textbf{1.182} \\
\bottomrule
\end{tabular}
\label{tab:primary}
\end{table*}
\section{Experiments}
\subsection{Experimental Setup}
\label{sec:ExpSettings}

We conduct experiments using data from the Chinese and US stock markets, including the CSI 300, CSI 500, S\&P 500, and NASDAQ 100 indices. For all datasets, we use basic technical indicators as input features. In backtesting, we predict the daily return ranking of each stock, select the top‑K stocks as positions, and compute their average true return as the daily investment return. The data are split chronologically into training (2018‑2022), validation (2023), and test (2024) sets to prevent data leakage.

We compare our method with state-of-the-art deep learning (DL) models (including LSTM \cite{hochreiter1997long}, GRU \cite{chung2014empirical}, Transformer \cite{vaswani2017attention}, PatchTST \cite{nie2022time}, DLinear \cite{zeng2023transformers}, iTransformer \cite{liu2023itransformer}, Crossformer \cite{zhang2023crossformer}, TimeMixer\cite{wang2023timemixer}, MASTER \cite{li2024master}), and UMI \cite{Yang2025LearningUM}, reinforcement learning (RL) models (including AlphaStock \cite{wang2019alphastock}, DeepTrader \cite{wang2021deeptrader}, and DeepPocket \cite{soleymani2021deep}), and GPTs for time series (including GPT4TS \cite{zhou2023one}, TIME-LLM \cite{jin2023time}, and aLLM4TS \cite{bian2024multi}).

We use the following six metrics for performance evaluation: Annualized Return Rate (ARR), Annualized Volatility (AVol), Maximum Drawdown (MDD), Sharpe Ratio (SR), Calmar Ratio (CR), and Information Ratio (IR). To eliminate fluctuations, we average the metrics over five repeated tests for each model. We utilize a 6-layer GPT-2 as the backbone network for prediction, thereby achieving a balance between performance and computational cost. Additionally,we apply the DB4 wavelet basis function as the initialization parameter.

\subsection{Financial Time Series Prediction}

\begin{figure}[tbp]
% \centering
    {\includegraphics[width=1\linewidth]{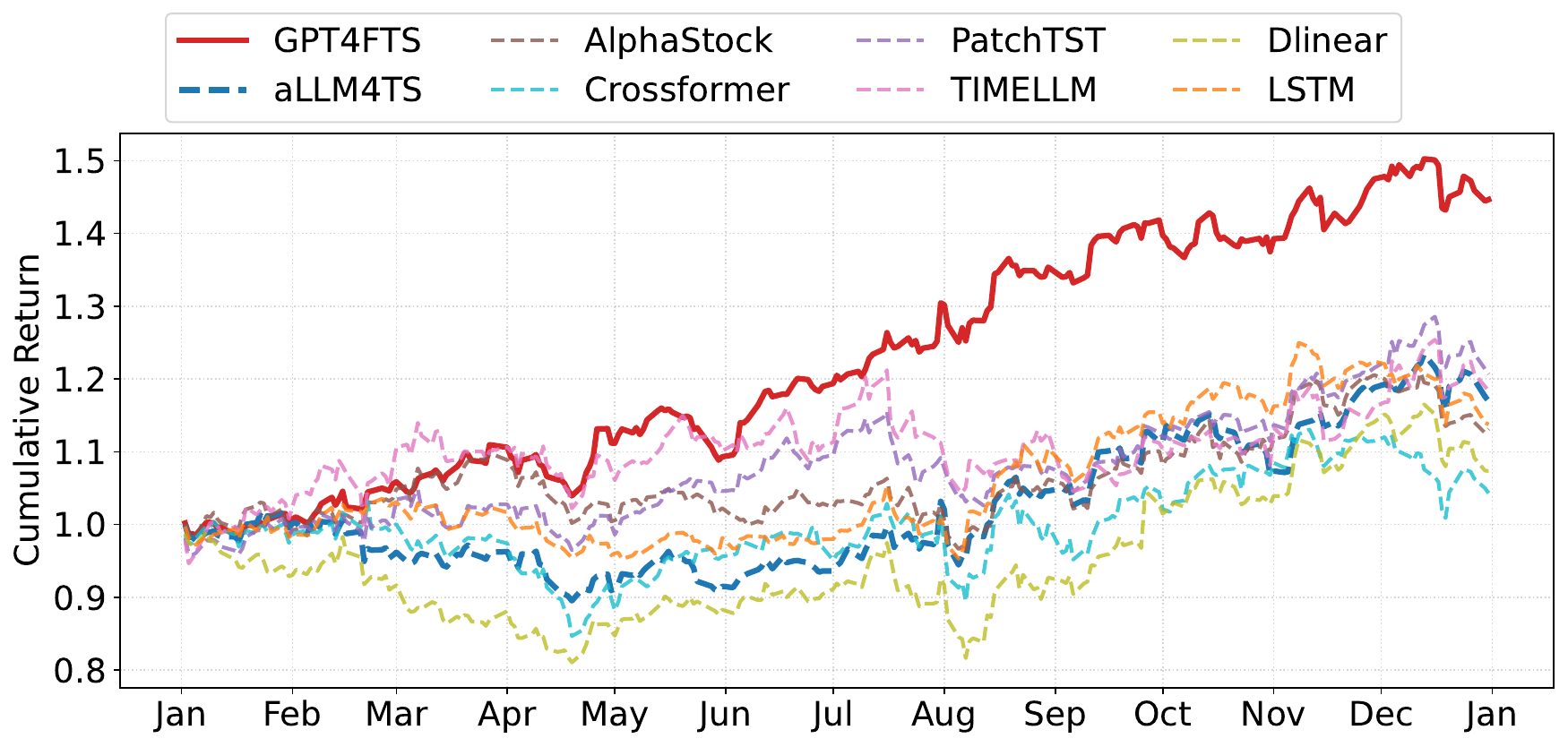}}
	\caption{The accumulated returns gained in the NASDAQ 100 dataset (2024) by  GPT4FTS and selected baselines.}
	\label{fig:port}
    \vspace{-1em}
\end{figure}

Table \ref{tab:port} compares GPT4FTS with baselines across datasets, showing it achieves the highest values in all four financial metrics—ARR, ASR,CR, and IR. Following financial research practices \cite{sawhney2021stock}, we use return-based metrics rather than error statistics to better capture practical trading performance (Section \ref{sec:ExpSettings}).

GPT4FTS delivers significant improvements. On CSI300, ARR is 0.528 (37.1\% higher than DeepTrader's 0.385), ASR is 2.262 (47.7\% above Crossformer's 1.532), and CR is 4.808 (64.4\% above AlphaStock's 2.924). On CSI500, ARR reaches 0.643 (14.2\% higher than TIME-LLM's 0.563) with an ASR of 1.829 (7.3\% improvement). For S\&P 500, ARR is 0.371 (15.6\% above GPT4TS's 0.321), ASR is 2.559 (25.8\% higher), and CR is 4.585 (5.5\% increase). The most notable gains are on NDX100: ARR of 0.446 (72.9\% above Transformer's 0.258), ASR of 2.772 (133\% higher than PatchTST's 1.190), CR of 7.066 (127\% above GPT4TS's 3.116), and the lowest Maximum Drawdown (0.063).

Table \ref{tab:port} includes traditional deep learning models (LSTM, GRU, Transformer), which show balanced but modest performance; specialized price-prediction models (AlphaStock, DeepPocket, etc.) that improve via reinforcement learning; general time-series SOTA methods (Dlinear, iTransformer, etc.) with comparable results; and large-model time-series approaches (GPT4TS, Time-LLM, aLLM4TS) that outperform most specialized models. GPT4FTS consistently surpasses all baselines, especially in risk-adjusted metrics, indicating superior return generation and risk management.

Figure \ref{fig:port} shows GPT4FTS's return curve in 2024 exceeds all baselines, with more stable performance. Its consistent superiority across markets highlights strong robustness and generalization for financial applications.

\subsection{Ablation} 
\begin{table}[tbp]
\small
\centering
\setlength{\tabcolsep}{2.5pt}
\caption{Performance evaluation of ablated models for portfolio management on four datasets.}
\begin{tabular}{ccccccccc}
\toprule
\multirow{2}{*}{Dataset} & \multirow{2}{*}{Metric} & \multicolumn{3}{c}{GPT-2} & \multicolumn{3}{c}{pt-GPT-2} \\
\cmidrule(lr){3-5} \cmidrule(lr){6-8}
& & Base & +SIPR & +Wave & Base & +SIPR & +Wave \\
\midrule
\multirow{2}{*}{CSI 300} & ARR & 0.070 & 0.177 & 0.185 & 0.312 & 0.417 & 0.416 \\
& ASR & 0.261 & 0.671 & 0.635 & 0.943 & 1.450 & 1.472 \\
\midrule
\multirow{2}{*}{CSI 500} & ARR & -0.120 & 0.125 & 0.156 & 0.376 & 0.529 & 0.434 \\
& ASR & -0.376 & 0.415 & 0.489 & 1.115 & 1.728 & 1.289 \\
\midrule
\multirow{2}{*}{NDX 100} & ARR & 0.161 & 0.179 & 0.224 & 0.183 & 0.267 & 0.343 \\
& ASR & 0.786 & 0.761 & 1.201 & 0.869 & 1.143 & 1.532 \\
\midrule
\multirow{2}{*}{S\&P 500} & ARR & 0.020 & 0.152 & 0.205 & 0.236 & 0.275 & 0.290 \\
& ASR & 0.084 & 1.040 & 1.272 & 1.481 & 1.854 & 1.897 \\
\bottomrule
\end{tabular}
\label{tab:ablation}
\vspace{-1em}
\end{table}
Comprehensive ablation studies thoroughly validate the effectiveness of each component in our framework, with detailed results summarized in Table~\ref{tab:ablation}. The significant performance degradation observed when using the original GPT-2 module underscores the critical role of our two-stage pre-training strategy, which is specifically designed to enable the model to more effectively capture the complex temporal dynamics inherent in financial time series data. Specifically, both the SIPR and wavelet modules consistently enhance prediction accuracy across all datasets when integrated with the pre-trained backbone, clearly demonstrating their complementary and synergistic contributions to the overall architecture. The SIPR module excels in capturing long-term dependencies through its advanced pattern matching mechanism, while the wavelet module strengthens multi-scale feature extraction by adaptively responding to the varying frequency characteristics present in market data. Notably, the wavelet component demonstrates its particular strength in handling volatile datasets, such as the NDX 100, where it significantly improves the ASR from 0.869 to 1.532 in the pt-GPT-2 configuration. These compelling results further emphasize the importance of customizing GPTs for financial prediction tasks through specialized architectural designs, and confirm that maintaining structural integrity in financial time series data is absolutely crucial for achieving optimal performance.

\subsection{Comparison with Fixed Wavelets}
\begin{figure}[tbp]
    {\includegraphics[width=1\linewidth]{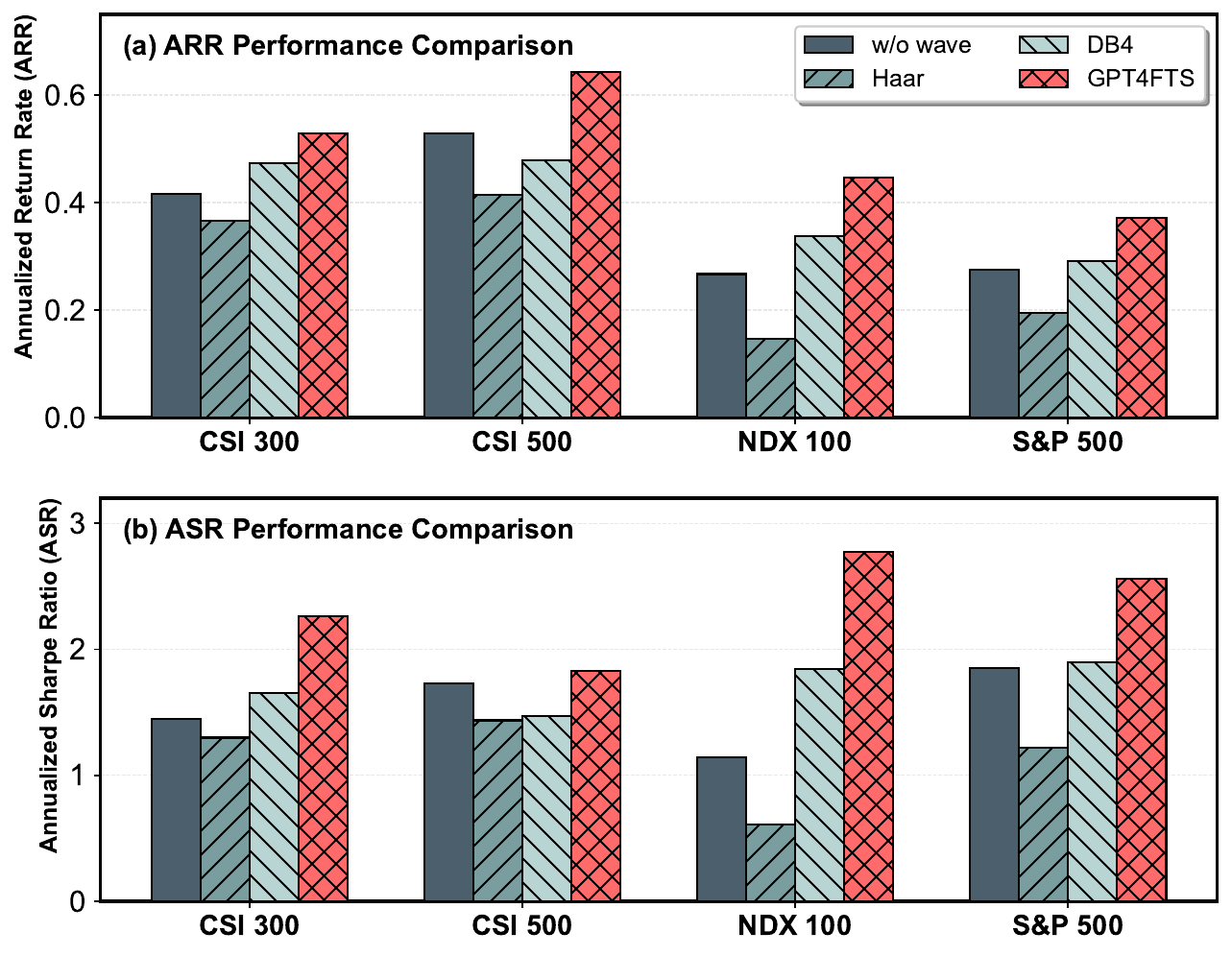}}
	\caption{Performance comparison between fixed and learnable wavelet transforms.}
	\label{tab:fixed_wavelets}
    \vspace{-1em}
\end{figure}

% \begin{table}[tbp]
% \small
% \centering
% \setlength{\tabcolsep}{5pt}
% \caption{Performance comparison between fixed and learnable wavelet transforms}
% \begin{tabular}{cccccccc}
% \toprule
% \multirow{2}{*}{Dataset} & \multirow{2}{*}{Metric} & \multicolumn{4}{c}{Method} \\
% \cmidrule(lr){3-6}
% & & w/o wave & Haar & DB4 & \textbf{GPT4FTS} \\
% \midrule
% \multirow{2}{*}{CSI 300} & ARR & 0.416 & 0.366 & 0.473 & \textbf{0.528} \\
% & ASR & 1.450 & 1.299 & 1.655 & \textbf{2.262} \\
% \midrule
% \multirow{2}{*}{CSI 500} & ARR & 0.529 & 0.414 & 0.479 & \textbf{0.643} \\
% & ASR & 1.728 & 1.436 & 1.471 & \textbf{1.829} \\
% \midrule
% \multirow{2}{*}{NDX 100} & ARR & 0.267 & 0.147 & 0.337 & \textbf{0.446} \\
% & ASR & 1.143 & 0.611 & 1.844 & \textbf{2.772} \\
% \midrule
% \multirow{2}{*}{S\&P 500} & ARR & 0.275 & 0.194 & 0.291 & \textbf{0.371} \\
% & ASR & 1.854 & 1.218 & 1.897 & \textbf{2.559} \\
% \bottomrule
% \end{tabular}
%  \vspace{-1em}
% \label{tab:fixed_wavelets}
% \end{table}
We conducted extensive experiments comparing our learnable wavelet filters with fixed wavelet transforms (Haar and Daubechies-4). The results in Fig~\ref{tab:fixed_wavelets} clearly demonstrate the superiority of our adaptive approach. While fixed wavelets provide modest improvements over the baseline, our learnable filters consistently achieve superior performance across all datasets. Notably, Haar wavelets exhibit significant performance degradation compared to Daubechies-4, highlighting the high sensitivity to wavelet type selection. These findings validate that our learnable wavelet framework offers significantly enhanced adaptability to diverse financial data distributions compared to traditional fixed wavelet transforms.

\subsection{Cross-Market Generalization}

To validate the generalization capability of our wavelet-based framework, we conducted cross-market transfer experiments. Models were trained on one market and directly tested on another without fine-tuning, providing a rigorous assessment of the learned representations' transferability.
The results in Table~\ref{tab:cross_market} demonstrate compelling generalization capabilities. Between Chinese markets, wavelet filters trained on CSI 500 achieve competitive performance on CSI 300, while CSI 300 filters transfer effectively to CSI 500. This indicates that the learned frequency-domain representations capture fundamental patterns transcending individual market characteristics. In US markets, filters show asymmetric transfer patterns. NDX 100 filters maintain reasonable performance on S\&P 500, though reverse transfer exhibits larger degradation. The consistent cross-market performance suggests our wavelet components learn universal multi-scale characteristics rather than market-specific artifacts. This generalization capability confirms the robustness of our approach and suggests practical utility for multi-market applications, where pre-trained wavelet filters could enable efficient knowledge transfer across different financial environments.

\begin{table}[tbp]
\small
\centering
\setlength{\tabcolsep}{7pt}
\caption{Cross-market generalization performance. Bold values indicate within-group best performance.}
\begin{tabular}{ccccccc}
\toprule
Target & Source & ARR & ASR & CR & IR \\
\midrule
\multirow{2}{*}{CSI 300} & CSI 300 & 0.528 & \textbf{2.262} & \textbf{4.808} & \textbf{1.965} \\
& CSI 500 & \textbf{0.536} & 1.791 & 3.575 & 1.647 \\
\midrule
\multirow{2}{*}{CSI 500} & CSI 500 & \textbf{0.643} & \textbf{1.829} & 2.839 & 1.591 \\
& CSI 300 & 0.598 & 1.815 & \textbf{3.439} & \textbf{1.628} \\
\midrule
\multirow{2}{*}{NDX 100} & NDX 100 & \textbf{0.446} & \textbf{2.772} & \textbf{7.066} & \textbf{2.371} \\
& S\&P 500 & 0.325 & 1.301 & 2.031 & 1.353 \\
\midrule
\multirow{2}{*}{S\&P 500} & S\&P 500 & \textbf{0.371} & \textbf{2.559} & \textbf{4.585} & \textbf{2.258} \\
& NDX 100 & 0.319 & 1.641 & 3.389 & 1.575 \\
\bottomrule
\end{tabular}
\label{tab:cross_market}
\end{table}

\subsection{Primary Evaluation Metrics}

While portfolio performance is important, it alone doesn't provide a full evaluation of a model's predictive capability. Therefore, we conducted additional experiments to assess the fundamental prediction accuracy of our method against strong baselines using MSE and MAE. As shown in Table~\ref{tab:primary}, GPT4FTS outperforms all baselines in terms of predictive accuracy across four financial datasets.

Specifically, on the challenging CSI 500 dataset, GPT4FTS achieves an MSE of 0.709$\times 10^{-3}$ and an MAE of 1.876$\times 10^{-3}$, outperforming Dlinear's 0.713$\times 10^{-3}$ and 1.887$\times 10^{-3}$, respectively. In US markets, on the S\&P 500, our method attains an MSE of 0.321$\times 10^{-3}$ and an MAE of 1.182$\times 10^{-3}$, surpassing aLLM4TS. 

Across diverse market conditions, including Chinese A-shares and US indices, our wavelet-enhanced framework demonstrates robust performance. Notably, on the technology-heavy NDX 100, GPT4FTS shows the greatest improvement, with an MSE of 0.461$\times 10^{-3}$. This confirms the state-of-the-art predictive accuracy of our method across multiple financial time series benchmarks.

\section{Conclusion}
This paper introduces GPT4FTS, a framework specifically designed for enhancing generative pre-trained transformers for financial time series prediction. The proposed architecture integrates a scale-invariant pattern recognition module with an adaptive dynamic patch segmentation strategy, enhanced by trainable wavelet transform operators for multi-resolution temporal dependency modeling. This combination enables systematic analysis of complex historical patterns in financial data through hierarchical feature extraction, effectively addressing the multi-scale characteristics of market dynamics. Comprehensive evaluations were conducted across four heterogeneous markets and diverse trading scenarios. Empirical results demonstrate the framework's superior performance against state-of-the-art baselines in both forecasting accuracy and generalization capability. To assess real-world performance, we integrated the model into the production-grade algorithmic trading infrastructure of a securities exchange platform. Comparative analysis with existing investment strategies reveals significant improvements in cumulative returns, confirming the operational efficacy and economic viability of our approach under real-market conditions.

\appendix

% \section*{Ethical Statement}

% There are no ethical issues.

% \section*{Acknowledgments}

% The preparation of these instructions and the \LaTeX{} and Bib\TeX{}
% files that implement them was supported by Schlumberger Palo Alto
% Research, AT\&T Bell Laboratories, and Morgan Kaufmann Publishers.
% Preparation of the Microsoft Word file was supported by IJCAI.  An
% early version of this document was created by Shirley Jowell and Peter
% F. Patel-Schneider.  It was subsequently modified by Jennifer
% Ballentine, Thomas Dean, Bernhard Nebel, Daniel Pagenstecher,
% Kurt Steinkraus, Toby Walsh, Carles Sierra, Marc Pujol-Gonzalez,
% Francisco Cruz-Mencia and Edith Elkind.

\bibliographystyle{named}
\bibliography{ijcai26}

@inproceedings{ariyo2014stock,
  title={Stock price prediction using the ARIMA model},
  author={Ariyo, Adebiyi A and Adewumi, Adewumi O and Ayo, Charles K},
  booktitle={UKSim-AMSS 16th International Conference on Computer Modelling and Simulation},
  pages={106--112},
  year={2014}
}

@inproceedings{wang2023timemixer,
  title={TimeMixer: Decomposable Multiscale Mixing for Time Series Forecasting},
  author={Wang, Shiyu and Wu, Haixu and Shi, Xiaoming and Hu, Tengge and Luo, Huakun and Ma, Lintao and Zhang, James Y and ZHOU, JUN},
  booktitle={International Conference on Learning Representations (ICLR)},
  year={2024}
}

@article{Yang2025LearningUM,
  title={Learning Universal Multi-level Market Irrationality Factors to Improve Stock Return Forecasting},
  author={Chen Yang and Jingyuan Wang and Xiaohan Jiang and Junjie Wu},
  journal={Proceedings of the 31st ACM SIGKDD Conference on Knowledge Discovery and Data Mining V.1},
  year={2025}
}

@inproceedings{chen2025simpletm,
title={Simple{TM}: A Simple Baseline for Multivariate Time Series Forecasting},
author={Hui Chen and Viet Luong and Lopamudra Mukherjee and Vikas Singh},
booktitle={The Thirteenth International Conference on Learning Representations},
year={2025}
}

@article{ghironi2006macroeconomic,
  title={Macroeconomic interdependence under incomplete markets},
  author={Ghironi, Fabio},
  journal={Journal of International Economics},
  pages={428--450},
  year={2006}
}

@article{frijns2010behavioral,
  title={Behavioral heterogeneity in the option market},
  author={Frijns, Bart and Lehnert, Thorsten and Zwinkels, Remco CJ},
  journal={Journal of Economic Dynamics and Control},
  pages={2273--2287},
  year={2010}
}

@inproceedings{Asgharian2002CrossSA,
  title={Cross Sectional Analysis of the Swedish Stock Market},
  author={Hossein Asgharian and Bj{\"o}rn Hansson},
  year={2002}
}

@inproceedings{machado2019lightgbm,
  title={LightGBM: An effective decision tree gradient boosting method to predict customer loyalty in the finance industry},
  author={Machado, Marcos Roberto and Karray, Salma and De Sousa, Ivaldo Tributino},
  booktitle={2019 14th International Conference on Computer Science \& Education (ICCSE)},
  pages={1111--1116},
  year={2019}
}

@article{taylor2016using,
  title={Using auto-regressive logit models to forecast the exceedance probability for financial risk management},
  author={Taylor, James W and Yu, Keming},
  journal={Journal of the Royal Statistical Society Series A: Statistics in Society},
  pages={1069--1092},
  year={2016},
}

@article{de2009predicting,
  title={Predicting the Brazilian stock market through neural networks and adaptive exponential smoothing methods},
  author={De Faria, EL and Albuquerque, Marcelo P and Gonzalez, JL and Cavalcante, JTP and Albuquerque, Marcio P},
  journal={Expert Systems with Applications},
  pages={12506--12509},
  year={2009}
}

@article{kim2003financial,
  title={Financial time series forecasting using support vector machines},
  author={Kim, Kyoung-jae},
  journal={Neurocomputing},
  pages={307--319},
  year={2003}
}

@article{wang2020forecasting,
  title={Forecasting method of stock market volatility in time series data based on mixed model of ARIMA and XGBoost},
  author={Wang, Yan and Guo, Yuankai},
  journal={China Communications},
  pages={205--221},
  year={2020}
}

@article{di2017recurrent,
  title={Recurrent Neural Networks Approach to the Financial Forecast of Google Asets},
  author={Di Persio, Luca and Honchar, Oleksandr and others},
  journal={International Journal of Mathematics and Computers in Simulation},
  pages={7--13},
  year={2017}
}

@inproceedings{huang2024generative,
  title={Generative Learning for Financial Time Series with Irregular and Scale-Invariant Patterns},
  author={Huang, Hongbin and Chen, Minghua and Qiao, Xiao},
  booktitle={International Conference on Learning Representations},
  year={2024}
}

@article{wu2021autoformer,
  title={Autoformer: Decomposition transformers with auto-correlation for long-term series forecasting},
  author={Wu, Haixu and Xu, Jiehui and Wang, Jianmin and Long, Mingsheng},
  journal={Neural Information Processing Systems},
  pages={22419--22430},
  year={2021}
}

@inproceedings{zhou2021informer,
  title={Informer: Beyond efficient transformer for long sequence time-series forecasting},
  author={Zhou, Haoyi and Zhang, Shanghang and Peng, Jieqi and Zhang, Shuai and Li, Jianxin and Xiong, Hui and Zhang, Wancai},
  booktitle={Proceedings of the AAAI conference on Artificial Intelligence},
  pages={11106--11115},
  year={2021}
}

@article{nie2022time,
  title     = {A Time Series is Worth 64 Words: Long-term Forecasting with Transformers},
  author    = {Nie, Yuqi and
               H. Nguyen, Nam and
               Sinthong, Phanwadee and 
               Kalagnanam, Jayant},
  booktitle = {International Conference on Learning Representations},
  year      = {2023}
}

@article{radford2019language,
  title={Language Models are Unsupervised Multitask Learners},
  author={Radford, Alec and Wu, Jeff and Child, Rewon and Luan, David and Amodei, Dario and Sutskever, Ilya},
  year={2019}
}

@article{brown2020language,
  title={Language Models are Few-shot Learners},
  author={Brown, Tom and Mann, Benjamin and Ryder, Nick and Subbiah, Melanie and Kaplan, Jared D and Dhariwal, Prafulla and Neelakantan, Arvind and Shyam, Pranav and Sastry, Girish and Askell, Amanda and others},
  journal={Neural Information Processing Systems},
  year={2020}
}

@article{touvron2023llama,
  title={Llama: Open and efficient foundation language models},
  author={Touvron, Hugo and Lavril, Thibaut and Izacard, Gautier and Martinet, Xavier and Lachaux, Marie-Anne and Lacroix, Timoth{\'e}e and Rozi{\`e}re, Baptiste and Goyal, Naman and Hambro, Eric and Azhar, Faisal and others},
  journal={arXiv preprint arXiv:2302.13971},
  year={2023}
}

@inproceedings{jin2024position,
  title={Position: What Can Large Language Models Tell Us about Time Series Analysis},
  author={Jin, Ming and Zhang, Yifan and Chen, Wei and Zhang, Kexin and Liang, Yuxuan and Yang, Bin and Wang, Jindong and Pan, Shirui and Wen, Qingsong},
  booktitle={International Conference on Machine Learning},
  year={2024}
}

@article{tang2024time,
  author       = {Hua Tang and
                  Chong Zhang and
                  Mingyu Jin and
                  Qinkai Yu and
                  Zhenting Wang and
                  Xiaobo Jin and
                  Yongfeng Zhang and
                  Mengnan Du},
  title        = {Time Series Forecasting with LLMs: Understanding and Enhancing Model Capabilities},
  journal      = {{SIGKDD} Explor.},
  pages        = {109--118},
  year         = {2024}
}

@article{wang2024stocktime,
  title={StockTime: A Time Series Specialized Large Language Model Architecture for Stock Price Prediction},
  author={Shengkun Wang and Taoran Ji and Linhan Wang and Yanshen Sun and Shang-Ching Liu and Amit Kumar and Chang-Tien Lu},
  journal={ArXiv},
  year={2024}
}

@article{yu2023temporal,
  title={Temporal Data Meets LLM--Explainable Financial Time Series Forecasting},
  author={Yu, Xinli and Chen, Zheng and Ling, Yuan and Dong, Shujing and Liu, Zongyi and Lu, Yanbin},
  journal={arXiv preprint arXiv:2306.11025},
  year={2023}
}

@article{yang2023fingpt,
  title={Fingpt: Open-source financial large language models},
  author={Yang, Hongyang and Liu, Xiao-Yang and Wang, Christina Dan},
  journal={arXiv preprint arXiv:2306.06031},
  year={2023}
}

@inproceedings{jin2023time,
  title={{Time-LLM}: Time series forecasting by reprogramming large language models},
  author={Jin, Ming and Wang, Shiyu and Ma, Lintao and Chu, Zhixuan and Zhang, James Y and Shi, Xiaoming and Chen, Pin-Yu and Liang, Yuxuan and Li, Yuan-Fang and Pan, Shirui and Wen, Qingsong},
  booktitle={International Conference on Learning Representations},
  year={2024}
}

@inproceedings{
    cao2024tempo,
    title={{TEMPO}: Prompt-based Generative Pre-trained Transformer for Time Series Forecasting},
    author={Defu Cao and Furong Jia and Sercan O Arik and Tomas Pfister and Yixiang Zheng and Wen Ye and Yan Liu},
    booktitle={International Conference on Learning Representations},
    year={2024}
}

@article{zhou2023one,
  title={One fits all: Power general time series analysis by pretrained lm},
  author={Zhou, Tian and Niu, Peisong and Sun, Liang and Jin, Rong and others},
  journal={Neural Information Processing Systems},
  pages={43322--43355},
  year={2023}
}

@article{bian2024multi,
    author={Bian, Yuxuan and Ju, Xuan and Li, Jiangtong and Xu, Zhijian and Cheng, Dawei and Xu, Qiang},
    title={Multi-Patch Prediction: Adapting LLMs for Time Series Representation Learning},
     journal={International Conference on Machine Learning},
    year={2024}
}

@inproceedings{liutimer,
  title={Timer: Generative Pre-trained Transformers Are Large Time Series Models},
  author={Liu, Yong and Zhang, Haoran and Li, Chenyu and Huang, Xiangdong and Wang, Jianmin and Long, Mingsheng},
  booktitle={International Conference on Machine Learning},
  year={2024}
}

@inproceedings{sawhney2021stock,
  title={Stock Selection via Spatiotemporal Hypergraph Attention Network: A Learning to Rank Approach},
  author={Sawhney, Ramit and Agarwal, Shivam and Wadhwa, Arnav and Derr, Tyler and Shah, Rajiv Ratn},
  booktitle={Proceedings of the AAAI Conference on Artificial Intelligence},
  volume={35},
  number={1},
  pages={497--504},
  year={2021}
}

@inproceedings{wavevit2022,
    title     = {Wave-ViT: Unifying Wavelet and Transformers for Visual Representation Learning},
    author    = {Yao, Ting and Pan, Yingwei and Li, Yehao and Ngo, Chong-Wah and Mei, Tao},
    booktitle = {Proceedings of the European conference on computer vision (ECCV)},
    year      = {2022},
}

@article{michau2022,
author = {Gabriel Michau  and Gaetan Frusque  and Olga Fink },
title = {Fully learnable deep wavelet transform for unsupervised monitoring of high-frequency time series},
journal = {Proceedings of the National Academy of Sciences},
volume = {119},
number = {8},
year = {2022},
doi = {10.1073/pnas.2106598119}
}

@article{liu2024moiraimoe,
  title={Moirai-MoE: Empowering Time Series Foundation Models with Sparse Mixture of Experts},
  author={Liu, Xu and Liu, Juncheng and Woo, Gerald and Aksu, Taha and Liang, Yuxuan and Zimmermann, Roger and Liu, Chenghao and Savarese, Silvio and Xiong, Caiming and Sahoo, Doyen},
  journal={arXiv preprint arXiv:2410.10469},
  year={2024}
}

@article{hochreiter1997long,
  title={Long Short-Term Memory},
  author={Sepp Hochreiter and J{\"u}rgen Schmidhuber},
  journal={Neural Computation},
  year={1997},
  pages={1735-1780}
}

@article{chung2014empirical,
  title={Empirical evaluation of gated recurrent neural networks on sequence modeling},
  author={Chung, Junyoung and Gulcehre, Caglar and Cho, KyungHyun and Bengio, Yoshua},
  journal={arXiv preprint arXiv:1412.3555},
  year={2014}
}

@article{vaswani2017attention,
 title={Attention is All you Need},
  author={Ashish Vaswani and Noam M. Shazeer and Niki Parmar and Jakob Uszkoreit and Llion Jones and Aidan N. Gomez and Lukasz Kaiser and Illia Polosukhin},
  booktitle={Neural Information Processing Systems},
  year={2017}
}

@inproceedings{wang2019alphastock,
  title={Alphastock: A buying-winners-and-selling-losers investment strategy using interpretable deep reinforcement attention networks},
  author={Wang, Jingyuan and Zhang, Yang and Tang, Ke and Wu, Junjie and Xiong, Zhang},
  booktitle={Proceedings of the 25th ACM SIGKDD international conference on knowledge discovery \& data mining},
  pages={1900--1908},
  year={2019}
}

@inproceedings{wang2021deeptrader,
  title={DeepTrader: a deep reinforcement learning approach for risk-return balanced portfolio management with market conditions Embedding},
  author={Wang, Zhicheng and Huang, Biwei and Tu, Shikui and Zhang, Kun and Xu, Lei},
  booktitle={Proceedings of the AAAI conference on artificial intelligence},
  volume={35},
  number={1},
  pages={643--650},
  year={2021}
}

@article{soleymani2021deep,
  title={Deep graph convolutional reinforcement learning for financial portfolio management--DeepPocket},
  author={Soleymani, Farzan and Paquet, Eric},
  journal={Expert Systems with Applications},
  volume={182},
  pages={115127},
  year={2021},
  publisher={Elsevier}
}

@inproceedings{li2024master,
  title={MASTER: Market-Guided Stock Transformer for Stock Price Forecasting},
  author={Li, Tong and Liu, Zhaoyang and Shen, Yanyan and Wang, Xue and Chen, Haokun and Huang, Sen},
  booktitle={Proceedings of the AAAI Conference on Artificial Intelligence},
  pages={162--170},
  year={2024}
}

@inproceedings{zeng2023transformers,
  title={Are transformers effective for time series forecasting?},
  author={Zeng, Ailing and Chen, Muxi and Zhang, Lei and Xu, Qiang},
  booktitle={Proceedings of the AAAI conference on artificial intelligence},
  pages={11121--11128},
  year={2023}
}

@article{liu2023itransformer,
  title={itransformer: Inverted transformers are effective for time series forecasting},
  author={Liu, Yong and Hu, Tengge and Zhang, Haoran and Wu, Haixu and Wang, Shiyu and Ma, Lintao and Long, Mingsheng},
  journal={arXiv preprint arXiv:2310.06625},
  year={2023}
}

@inproceedings{zhang2023crossformer,
  title={Crossformer: Transformer utilizing cross-dimension dependency for multivariate time series forecasting},
  author={Zhang, Yunhao and Yan, Junchi},
  booktitle={International Conference on Learning Representations},
  year={2023}
}

@article{zhu2023minigpt,
  title={Minigpt-4: Enhancing vision-language understanding with advanced large language models},
  author={Zhu, Deyao and Chen, Jun and Shen, Xiaoqian and Li, Xiang and Elhoseiny, Mohamed},
  journal={arXiv preprint arXiv:2304.10592},
  year={2023}
}

\end{document}